\documentclass[letterpaper, 10 pt, conference]{ieeeconf}  % Comment this line out if you need a4paper

\IEEEoverridecommandlockouts                              % This command is only needed if 
\usepackage{graphics} % for pdf, bitmapped graphics files
\usepackage{epsfig} % for postscript graphics files
\usepackage{mathptmx} % assumes new font selection scheme installed
\usepackage{times} % assumes new font selection scheme installed
\usepackage{amsmath} % assumes amsmath package installed
\usepackage{amssymb}  % assumes amsmath package installed
\usepackage[linesnumbered, ruled]{algorithm2e}

  \usepackage[switch]{lineno}
\usepackage{booktabs}
\usepackage{threeparttable}
\usepackage{romannum}
\usepackage{soul}

\usepackage{color, xcolor}
\usepackage{here}
\usepackage{graphicx}
\usepackage{subfigure}
\usepackage{cite}
\usepackage{bm}
\usepackage{float}
\usepackage{mathtools}
\usepackage{amsmath,amsfonts}

\title{\LARGE \bf
Evolutionary-Based Online Motion Planning Framework for \\Quadruped Robot Jumping
}

\author{Linzhu Yue$^{1}$, Zhitao Song$^{1}$, Hongbo Zhang$^{1}$, Xuanqi Zeng$^{1}$, Lingwei Zhang$^{2}$, and Yun-Hui Liu$^{1}$ % <-this % stops a space
\thanks{$^{1}$ L. Z. Yue, Z. T. Song, H. B. Zhang, X. Q. Zeng, and Y.-H. Liu are with the Department of Mechanical and Automation Engineering, The Chinese University of Hong Kong.
        {\tt\small lzyue@mae.cuhk.edu.hk}}%
\thanks{$^{2}$ L. W. Zhang is with the Hong Kong Centre for Logistics Robotics. 
        }%
\thanks{* Corresponding author: Y.-H. Liu  
        {\tt\small yhliu@cuhk.edu.hk}}
\thanks{The InnoHK Clusters support this work via the Hong Kong Centre of Logistics Robotics.}% <-this % stops a space
}

\begin{document}

\maketitle
\thispagestyle{empty}
\pagestyle{empty}

%%%%%%%%%%%%%%%%%%%%%%%%%%%%%%%%%%%%%%%%%%%%%%%%%%%%%%%%%%%%%%%%%%%%%%%%%%%%%%%%
\begin{abstract}

Offline evolutionary-based methodologies have supplied a successful motion planning framework for the quadrupedal jump. However, the time-consuming computation caused by massive population evolution in offline evolutionary-based jumping framework significantly limits the popularity in the quadrupedal field. This paper presents a time-friendly online motion planning framework based on meta-heuristic Differential evolution (DE), Latin hypercube sampling, and  Configuration space (DLC). The DLC framework establishes a multidimensional optimization problem leveraging centroidal dynamics to determine the ideal trajectory of the center of mass (CoM) and ground reaction forces (GRFs). The configuration space is introduced to the evolutionary optimization in order to condense the searching region. Latin hypercube sampling offers more uniform initial populations of DE under limited sampling points, accelerating away from a local minimum. This research also constructs a collection of pre-motion trajectories as a warm start when the objective state is in the neighborhood of the pre-motion state to drastically reduce the solving time. The proposed methodology is successfully validated via real robot experiments for online jumping trajectory optimization with different jumping motions (e.g., ordinary jumping, flipping, and spinning).
% (Video$^\star$)

\end{abstract}

%%%%%%%%%%%%%%%%%%%%%%%%%%%%%%%%%%%%%%%%%%%%%%%%%%%%%%%%%%%%%%%%%%%%%%%%%%%%%%%%
\section{INTRODUCTION}

A crucial aspect of a quadrupedal robot's capacity to traverse various terrains is its ability to perform jumping motions in tough natural surroundings. To adapt to uneven terrains, many researchers have focused on locomotion via diverse gaits (e.g., bounding, walking). Some works have already yielded outstanding results, such as high-speed bounding in \cite{Di_01} and \cite{Park_02}. However, crossing over unavoidable obstacles (e.g., deep canals and roadblocks) usually requires a robust and high-performance jumping motion controller.
A core difficulty for quadruped jumping is generating trajectories in real time under kino-dynamic constraints \cite{Donald_03} (e.g., physical constraints). \cite{Chignoli_04} and \cite{Bellegarda_05} have already achieved unforgettable results. However, the laborious calculation of offline trajectories makes it difficult or impossible to apply to jobs involving frequent re-planning. This motivates the development of a unified framework that satisfies online planning.
\vspace{-0.2cm}
\begin{center}
\begin{figure}[!ht]
\centering
\includegraphics[width=3.2in,height=3.2in]{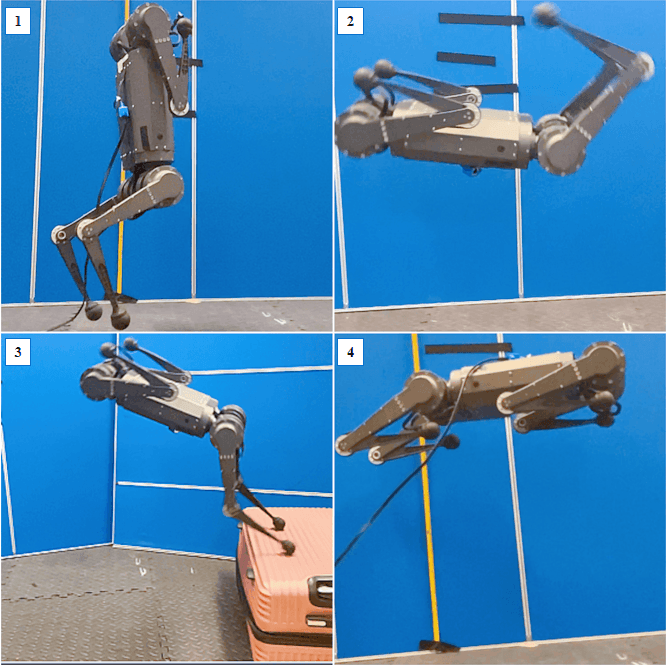}
\caption{ Various jumping motion experiments to validate the proposed online motion framework. (1) Two-leg vertical jumps with $\theta =\frac{\pi}{2}$. (2) Back-flip jumps with $\theta =-2\pi$. (3) Back-flip jumps from 0.3 (m) height platform (4) Four-leg vertical jumps reach the max height of 0.7 (m). }\label{problemIllustration}
\vspace{-0.3cm}
\end{figure}
\vspace{-0.7cm}
\end{center}
\quad Some publications address jumping trajectory optimization issues subject to complex kino-dynamics constraints using Reinforcement Learning (RL). The RL approach has shown a remarkable capacity for complicated locomotion on legged robots\cite{Lee_06, Li_07, Choi_08}. Recently, some works have used RL to deal with the jumping of quadruped robots. Learned policies, inspired by cat landing behavior, were used to control the robot's posture in the landing phases.\cite{Rudin_07}. However, few works are focusing on planning multiple complicated jumping motions using a single policy.

Gradient-based trajectory optimization is a commonly employed optimization method in robot jumping control. The MIT research uses gradient-based optimization algorithms to assist the robots Cheetah 3 jump on a high desk (0.76 (m)) and Mini-Cheetah to cover a variety of jumping motions with an online 3-D jumping trajectory optimization approach\cite{Nguyen_03, Chignoli_15}, respectively. Similar to \cite{Gilroy_13}, they use collocation-based optimization to build offline trajectories over dynamically-feasible barriers. Additionally, \cite{Ding_14} utilized a mixed-integer convex program to circumvent the reference motion limits; however, this method must be optimized offline.

Heuristic algorithms can efficiently solve optimization problems with complex constraints, which supplies a new approach for generating jumping motions. Differential Evolution (DE) is a heuristic-based algorithm proposed by Storn and Price \cite{Storn_DE_1005}. DE algorithms have been utilized in robotics, signal processing, and other industries to address complicated optimization problems \cite{Deng_2021}. In our previous work, DE is employed to generate offline jumping trajectories for a quadrupedal robot \cite{Zt_2022}. However, this offline approach is inherently inferior for systems with online re-planning requirements. Moreover, the technique called Latin hypercube sampling (LHS) can be used to generate an initialization population for DE, increasing the convergence speed in low-dimensional space (typically less than 20 dimensions)\cite{Wang_2022, Dong_2021}.

In our work, we try to accelerate the DE algorithm through three techniques: search space compressing without losing the jumping motion performance, careful selection of the initial population, and a warm start by pre-calculation. By adding the conditioned configuration space of the robot, the smaller searching space enables the DE approach to reduce the population and the number of iterations, hence accelerating the optimization time.
The LHS gives more uniform initial populations of the DE algorithm, which accelerates away from the local minimum. Also, when the desired state is in close proximity to solved states saved in the pre-motion library, pre-calculated optimization variables can be shared with new evolution as a warm start. 
To sum up, the DE algorithm based on C-space, LHS, optimization variables transformation, and Pre-motion library is used to construct the proposed framework. 

In this work, we intend to answer the following questions:  a) How to design a time-friendly optimization framework for online motion planning using an evolution-based technique? b) How do the configuration space, Latin hypercube sampling, and optimization variables transformation accelerate the convergence speed of the optimization? c) How to produce a series of trajectories for the Pre-motion Library? 
% as the warm-start to assist in removing the time-consuming and heavy online computation power limitation of conventional methods?
% the initial population of the differential evolution method has a significant impact on the convergence of the fitness function; the Latin hypercube sampling (LHS) used to generate a random global initialization population increases the convergence speed in low-dimensional space (typically less than 20 dimensions)\cite{Wang_2022, Dong_2021}.

% \subsection{Contribution}
Our primary contributions can be shown as follows:
\begin{enumerate}
\item A time-friendly online motion planning framework for quadruped jumping based on the meta-heuristic Differential evolution, Latin hypercube sampling, and  Configuration space (DLC) algorithm is proposed, which can generate various jumping trajectories online.
\item We creatively combine configuration space, Latin hypercube sampling, and the Pre-motion Library to reduce optimization time.
\item The algorithm has been verified online by various jumps on a real quadruped robot (see Fig. \ref{problemIllustration}).
\end{enumerate}

Moreover, the current study differs from our prior work \cite{Zt_2022} in that it only addresses how to solve a quadruped robot jumping problem by an evolutionary algorithm but does not consider how to overcome the time-consuming restriction.
% The remainder of this paper is structured as follows: Section \Romannum{2} introduces the robot reduce-order model and shows the centroidal dynamics of our jumping algorithm. Section \Romannum{3}, We detail how the bi-level online trajectory optimization framework was established, including how optimization formulation, optimization variables, and C-space were produced. Section \Romannum{4} details the experiments and software implementation. Finally, Section \Romannum{5} concludes by summarizing the paper and outlining the objective of the following stage.

\section{Models and dynamics}
\begin{figure}[htb]
  \centering
% \hspace{-2mm}
 \subfigure[]{
    \label{CaRe1} %% label for second subfigure
    \includegraphics[width=0.25\textwidth]{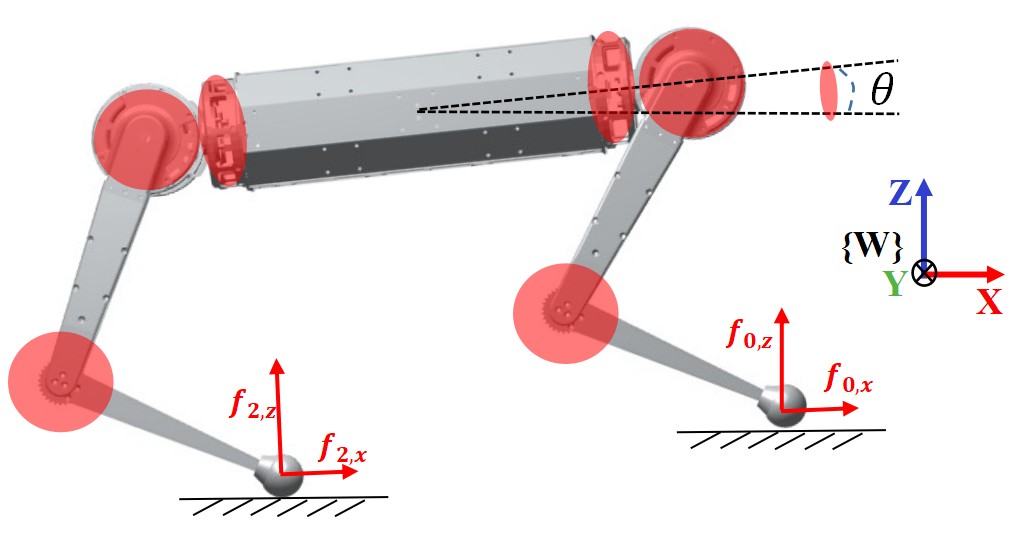}}
  \subfigure[]{
    \label{CaPe2} %% label for first subfigure
    \includegraphics[width=0.2\textwidth]{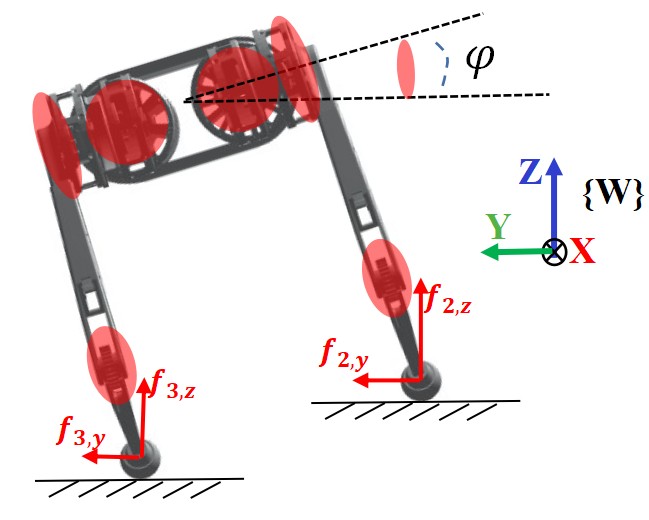}}
\vspace{-3mm}
  \caption{2D planar models schematic diagram used in DLC framework. (a) and (b) represent the sagittal plane and coronal plane, respectively. The red arrow indicates the GRFs of different jumping planes. The red area indicates that the 2D planar model reduces the 18 DoFs for diverse jump motions to 7 DoFs.}
  \label{2d_planar_model} %% label for entire figure
\end{figure}

The objective of this section is to present the centroidal dynamics and 2D planar model for the DLC framework. The reduced-order dynamic model of jumping motion treats the robot as a single rigid body (SRB) with a specified moment of inertia for the optimization process. 
% The robot's torso and legs are treated as one unit because of the leg's obscenely slight inertia in comparison to the body. 
Moreover, this work considers the 2D planar model (i.e., sagittal and coronal planes) to the framework as shown in Fig. \ref{robot_coordinate} and Fig. \ref{2d_planar_model}. The $\bm x$ presents the system state.
\begin{subequations}
\begin{eqnarray}
& \bm x:=\lbrack {\bm P_C^T} \quad {\bm \Theta^T}\quad {\bm V_C^T} \quad  ^{B}{\bm \omega^T}\rbrack^T \in \mathbb{R}^{12}\label{q_joint_2}\\
&{\bm Q}:=[{\bm q_{i}} \quad {\dot{\bm q}_{i}}] \in \mathbb{R}^{24} \label{q_joint_1}
\end{eqnarray}
\end{subequations}
where ${\bm P_C} \in \mathbb{R}^{3}$ is the position of the robot center of mass (CoM) w.r.t inertial frame; ${\bm \Theta \in \mathbb{R}^{3}}$ represents the Euler angles of the robot; ${\bm V_C \in \mathbb{R}^{3}}$ is the velocity of the CoM. $^{B}{\bm \omega} \in \mathbb{R}^{3}$ is the angular velocity of CoM represented in the robot frame $B$. $\bm q_i \in \mathbb{R}^3$ and $\bm \dot{\bm q}_i \in \mathbb{R}^3$ are the joint angles and velocities of each leg. $i$ is the number of feet.
The GRFs ${\bm u}:=[{\bm f_{i}}]  \in \mathbb{R}^{12},{\bm f_{i}} \in \mathbb{R}^3$ is the dynamic system control input at each contact point acquired by optimization. ${\bm r_i}$ is the vector from CoM to the robot foot. Then the body net wrench $\mathcal{F} \in \mathbb{R}^{6}$ of the CoM is shown as follows:
\vspace{-0.1cm}
\begin{eqnarray}
\mathcal{F}=\left[\begin{array}{c}
\boldsymbol{F}_{\textrm{c}} \\
\boldsymbol{\tau}_{\textrm{c}}
\end{array}\right]=\sum_{i=1}^4\left[\begin{array}{l}
\boldsymbol{f}_i \\
\boldsymbol{r}_i \times \boldsymbol{f}_i
\end{array}\right], 
\end{eqnarray}
where $\boldsymbol{F}_{\textrm{c}}$ and $\boldsymbol{\tau}_\textrm{{c}}$ represent the total force and torque of CoM.
Moreover, the simplified model for jumping motions decreases the 18 Degrees-of-Freedom (DoFs) to 7 (including 6 leg joints and an angle of the jumping plane, see Fig. \ref{2d_planar_model}).

Then the equations of the centroidal dynamics model\cite{tedrake_10_5} are given in \eqref{centroidal_model} along with the coordinates defined in Fig.~\ref{robot_coordinate}.
\begin{subequations}
% \begin{eqnarray}
\begin{align}
\ddot{\bm P}_C(t)&=\frac{\sum_{i=1}^4 \boldsymbol{f}_i}{m}- {\bm g} \label{acc}\\
\frac{\mathrm{d} ({\bm I \bm \omega})}{\mathrm{d} t}&=\bm{\tau}_{\textrm{c}}+\bm{0}_{3 \times 1}\times(\mathrm{m} \cdot{ } \bm{g})
\label{rcc},
\end{align}
\label{centroidal_model}
% \end{eqnarray}
\end{subequations}
where ${\bm g} \in\mathbb{R}^3$ represents gravitational acceleration. ${^B\bm{I} } \in \mathbb{R}^{3\times3}$ is the robot's rotation inertial tensor which is assumed as a constant in this work, $\textrm{diag}(^B\bm{I})=[0.07,0.26,0.242]^T$.
In addition, our framework classifies jumping motions into four phases: four-foot contact, two-foot contact, flying phase, and landing phase.
\begin{center}
\vspace{-0.5cm}
\begin{figure}[htb]
\centering
\includegraphics[width=3in]{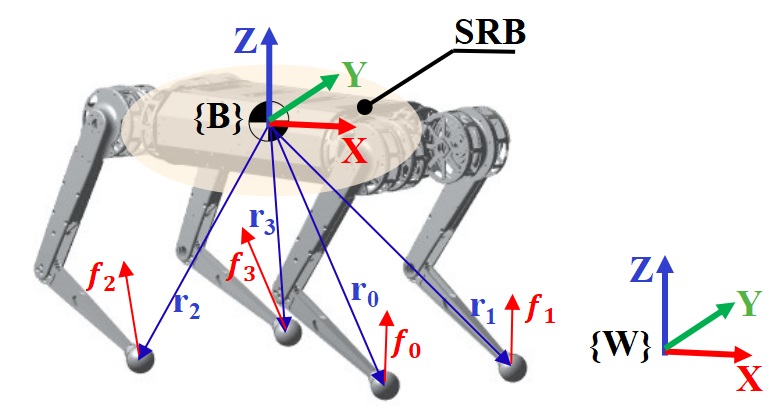}
\caption{A model of a single rigid body (SRB) utilized in the framework for optimization. The blue arrow represents the CoM to the plantar position vector, while the red arrow represents the Ground Reaction Forces (GRFs).}\label{robot_coordinate}
\end{figure}
\vspace{-0.8cm}
\end{center}

\begin{figure*}[h]
\centering
% \vspace{-0.8cm}
\includegraphics[width=5.5in]{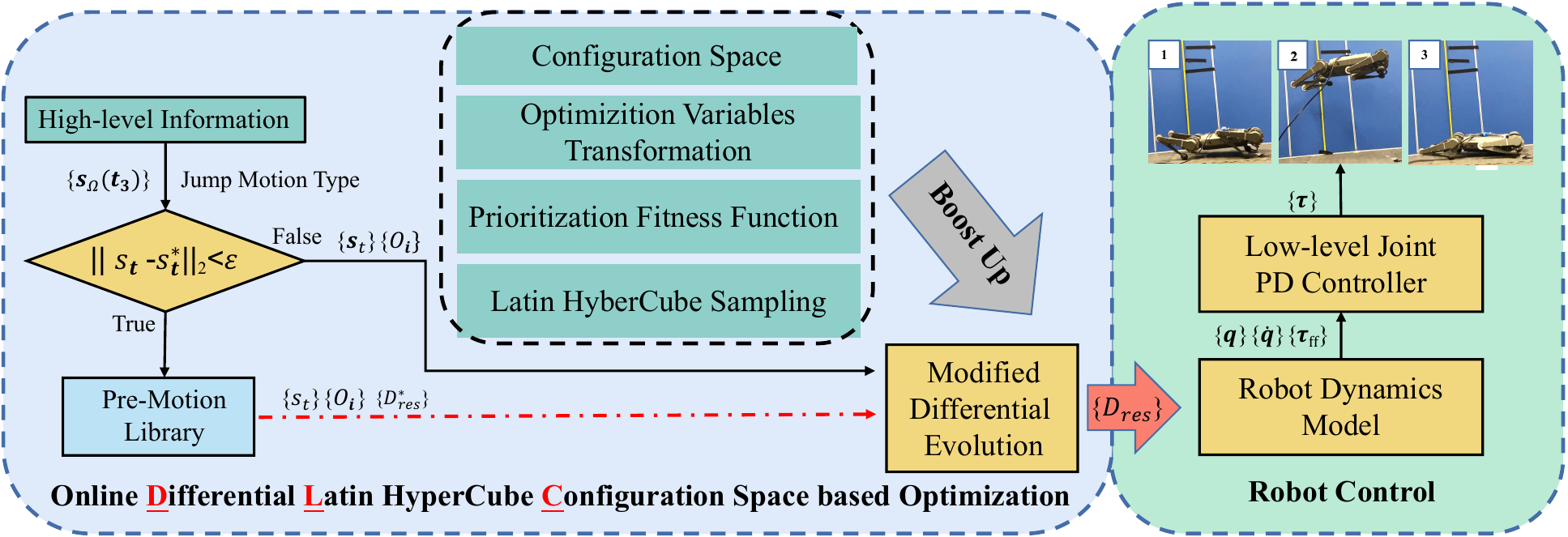}
\vspace{-0.2cm}
% \captionsetup{justification=centering}
\caption{Overview of Online DLC optimization jumping framework, based on LHS, DE, and C-space. The red dot line means using the trajectory from the Pre-motion Library. The motion planning procedure is shown by the blue blocks. The low-level controller is shown by the green blocks.}
\label{jump_framework}
\vspace{-0.5cm}
\end{figure*}
\section{Jumping Motion Planning Framework}
To design a time-friendly optimization framework for online motion planning, we improve the meta-heuristic \textbf{D}ifferential evolution algorithm by introducing the \textbf{L}atin Sampling algorithm to replace the random initial population and the \textbf{C}onfiguration space to further reduce the optimization algorithm's search domain.
Additionally, the pre-motion library is also used as the warm-start. 
% We provide a fitness function with manually determined constraint priority to enable the DE to quickly optimize complex jumping motions. 
% This generates the method to converge on the highest efficiency, ensuring online optimization.
% Those innovations increase DE algorithm convergence speed compared to previous evolutionary-based algorithms. It satisfies the online optimization requirements for the quadrupedal jump.
% \vspace{-1.2cm}
% To achieve the goal of designing a time-friendly optimization framework
% for online motion planning, we improve the meta-heuristic \textbf{D}ifferential evolution algorithm by introducing the \textbf{L}atin Sampling algorithm to make it uniformly sampled to lower the random sampling time, while the \textbf{C}onfiguration space is novelty added to the search space to further reduce the search domain of the optimization algorithm. To enable the DE to rapidly optimize complex jumping motions, we introduce a fitness function with manually defined constraint priority, which generates the whole algorithm to converge to the highest efficiency, achieving the goal of online optimization.
% After the three significant enhancements in our framework, the convergence speed DE Algorithm is substantially enhanced compared with other common evolutionary-based algorithms. It can meet the online optimization requirements in different robot jumping motions.
% \vspace{-0.1cm}
\subsection{Optimization Formulation \label{op_formulation}}
The objective of this section is to build the optimization problem and optimization objectives. Additionally, unlike the gradient-based method, our evolutionary-based optimization framework's cost function is a well-designed priority hierarchical fitness function.
% \vspace{-0.6cm}
\begin{eqnarray}
\begin{aligned}
& \underset{D_{\textrm{opt}}}{\text{minimize}}
& & 10^{L}-\sum\limits_{n=3}^{L}(10^{n-3}\sigma_{n}W_{n})+ W_{1}\zeta \\
& \text{subject to}
& & \bm{x}\left(k+1\right)=\bm{x}\left(k\right)+{\Delta t}\dot{\bm{x}}\left(k\right) \\
&&& \bm \dot{\bm x}_{k+1}=g(\bm u_k,\bm x_k)\\
&&&\boldsymbol{x}_{k} \in \mathbb{X}, k=1,2, \cdots, N \\
&&&\boldsymbol{u}_{k} \in \mathbb{U}, k=1,2, \cdots, N\\
&&&\boldsymbol{q}_{k} \in \mathbb{Q}, k=1,2, \cdots, N\\
&&&\bm{x}\left(0\right)=\bm{x}_{0},
\bm{x}\left(N\right)=\bm{x}_{\textrm{target}}
\end{aligned}
\label{optimization_formula}
\end{eqnarray}
where $\zeta=\int_{0}^T(|\bm{\tau}(t)\bm{\dot{q}}(t)|)dt$ is the energy consumption of the motion. $D_{\textrm{opt}}$ is the optimization variables whose meaning and selection are elaborated in the next section.
${\sigma}_n \in \mathbb{R}$ is the differences between the system state produced by the evolutionary algorithm optimization iteration process and the feasible set specified by C-space. $W_n \in [0,1] $ is a weight that indicates the significance of one constraint to the optimization problem. $N$ is the evolution population number (ref to Algorithm \ref{algo_de}); $L \in \mathbb{R}$ is the total number of layering priority constraints. $\mathbb{X}$, $\mathbb{U}$ and $\mathbb{Q}$ are the feasible sets according to kino-dynamics constraints. $\bm{x}\left(0\right)$ and $\bm{x}\left(N\right)$ are the initial state and the desired state for the robot; $\bm \dot{\bm x}_{k+1}=g(\bm u_k,\bm x_k)$ represents the centroidal dynamics combination form with respect to (\ref{acc}) and (\ref{rcc}).

\subsection{Optimization Parameters and Transformation}
The objective of this section is to design optimization variables for the DLC framework of different jump motions.
We utilize the system state as the optimization variable instead of employing polynomial parameters. 

Here, the scenario of front jumping in the sagittal plane illustrates how to generate the optimization variables and do the optimization variables transformation. \\
\textbf{Assumption 1}: The force along the y-axis is zero.\\
\textbf{Assumption 2}: Leg 0 and leg 1 has the equivalent force, the same as the rear two legs. That is, $\bm f_0=\bm f_1$ and $\bm f_2=\bm f_3$.\\
\textbf{Assumption 3}: The x-axis force of the front and back feet are equal when ${t} \in [0,  {t}_1]$. %($t \in [0,  {t}_1]$)
Based on the assumptions, the equation of GRFs of the front jump can be simplified as follows:
\begin{subequations}
\begin{align}
\bm {f}_i&=\left\{
\begin{array}{rcl}
a_1t+a_0 & & {{t} \in [0,  {t}_1]}\\
b_2t^2+ b_1t+b_0 & & {{t} \in \left[{t}_1,{t}_2 \right]}\\
0 & & {{t} \in \left [{t}_2,{t}_3\right ]}
\end{array} \right. \label{coefficient_front},\\
\Lambda&=[a_0,a_1,b_0,b_1,b_2]
\label{coefficient_d},
\end{align}

\end{subequations}
By using a 2D simplified model of front jumping motion, the robot state can be represented as $\bm s_\Omega(t) = [x_c,z_c,\theta]$, where $[x_c,z_c]$ are the position of CoM and $\theta$ is the pitch angle. We can get the analytical expression of ${\bm s_\Omega(t)}$ from GRFs given in (\ref{coefficient_front}) using the centroidal dynamics models. In addition, there are 12 polynomial coefficients for one jump motion according to assumption 2. The robot's state should then ideally be utilized for optimization. Furthermore, to easily bound the $\Lambda$, we convert polynomial coefficients into expressions based on $\bm s_\Omega(t)$.  we choose robot states under four time points together with three durations of different jumping phases ($[\frac{t_1}{2},t_1,t_2,t_3]$). 
Then, we can select the optimization variables given in (\ref{d_opt}).
% First, we choose four time points ($[\frac{t_1}{2},t_1,t_2,t_3]$) to obtain 12 positions and orientations w.r.t the CoM in (\ref{coefficient_12}). Then we construct 12 equations based on the dynamics of centroidal dynamics (\ref{acc}) and (\ref{rcc}) with using 12 polynomial coefficients. 
% \begin{eqnarray}
% \bm{s}_\Omega^{\textrm{total}}:=[\bm{s}_\Omega(\frac{t_1}{2}),\bm{s}_\Omega(t_1),\bm{s}_\Omega(t_2),\bm{s}_\Omega(t_3)] \in\mathbb{R}^{12}
% \label{coefficient_12},
% \end{eqnarray}
% Second, we can obtain the (\ref{coefficient_d}) with (\ref{coefficient_12}) by solving  12 equations. Then, we transform polynomial parameters with no physical meaning into formulations with physical meaning. Once the polynomial parameters are determined, we can determine the GRFs to produce trajectories w.r.t the CoM.
% Finally, the $\bm{s}_\Omega(t_3)$ will be obtained from high-level information, the remaining 12 variables are unknown and must be determined by optimization in (\ref{d_opt}). These three steps also apply to other jumping motions.
\begin{eqnarray}
\bm{D}_{\textrm{opt}}:=[\bm{s}_\Omega(\frac{t_1}{2}),\bm{s}_\Omega(t_1),\bm{s}_\Omega(t_2),t_{opt}]^T \in \mathbb{R}^{12}, \label{d_opt}
\end{eqnarray}

\begin{algorithm} 
% \vspace{-0.8cm}
\caption{DLC Algorithm}
\label{algo_de} 
\SetKwInOut{Input}{input}\SetKwInOut{Output}{output}
	\Input{$ \bm s_t,\bm O_k, {\bm D}_{\textrm{res}}^*,{\textrm{Maxgen}},NP,\bm{D}_{\textrm{opt}},r,\varepsilon$}
 
	\Output{${\bm D_{\textrm{res}}}$}
	 \BlankLine 
  	 \emph{$g\leftarrow 1, \bm k \leftarrow [{\bm s_t},{\bm O_k}],{\bm {s}_m \in \mathbb{R}^{12} \leftarrow \bm{D}_{\textrm{opt}}}$ }\;
  	 
	 \emph{$\bm {\Omega_s} \leftarrow \{\bm {s}_m \ \arrowvert \ s_{m,1 \sim 9} \in {\bm \Omega_{C}},s_{m,10 \sim 12} \in {\bm \Omega_{T}}\}$ }\; 
	 
     \emph{$\bm{\Omega}_s^* \leftarrow \{\bm{s}_m \in \bm{\Omega}_s \ \arrowvert \ \Arrowvert \bm{s}_m - {\bm D_{\textrm{res}}^*} \Arrowvert < {r} \} $}\; 
 	 
  	\uIf {$\Arrowvert \bm{s}_t - {\bm s_{t}^*} \Arrowvert_2 < \varepsilon$} {
        \emph{$\bm {s}_m(g)\leftarrow LHS({\bm \Omega_s}, NP)$}\;
 		
    } \Else{
        \emph{$\bm {s}_m(g)\leftarrow LHS({\bm \Omega^*_s}, NP)$}\;
    }
	 \While{Fitness($\bm D_{\textrm{res}}(g) , {\bm k})> {\varepsilon} \ \bm{or}\ g < \rm {Maxgen}$} { 
	 	\For{$m\leftarrow 1$ \KwTo $NP$}{
	 	\emph{Mutation and Crossover}\; 
	 	\For{$n\leftarrow 1$ \KwTo $12$}{
            $v_{m,n}(g)\leftarrow M(s_{m,n}(g))$\;
            $u_{m,n}(g)\leftarrow C(s_{m,n}(g),v_{m,n}(g))$\;
 	 	   }
 	 	   \emph{Selection}\;
 	 	\uIf {Fitness(${\bm U}_m(g),{\bm k}$)$<$Fitness(${\bm s}_m(g),{\bm k}$)} {
            \emph{${\bm s}_m(g)\leftarrow {\bm U_m(g)}$}\;
 			\lIf{Fitness(${\bm s}_m(g),{\bm k}$)$<$Fitness(${\bm D}_{\textrm{res}}(g),{\bm k}$)}{$\bm D_{\textrm{opt}}\leftarrow {\bm s}_m(g)$
 			}
 			
        } \Else{
            ${\bm s_m(g)}\leftarrow \bm s_m(g)$\;
        }
  }
   \emph{$g \leftarrow g+1$}\;} 
   
    \end{algorithm}
    % \vspace{-0.4cm}
 \DecMargin{1em} 
 \vspace{-0.2cm}
\subsection{C-space and Kino-dynamic Constraints}
\begin{figure}[!h]
  \centering
 \subfigure[]{
    \label{CaRe} %% label for second subfigure
    \includegraphics[width=1.5in]{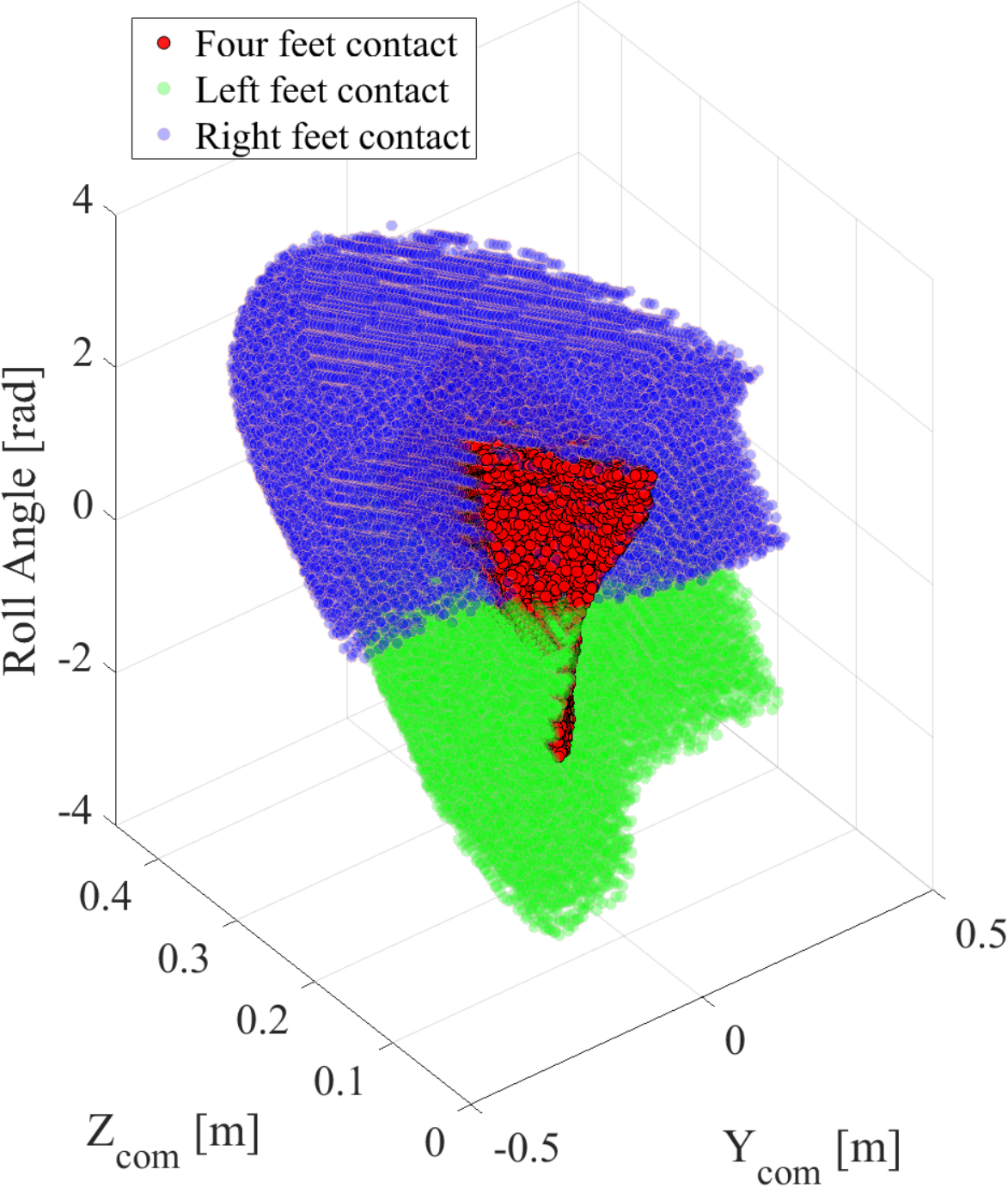}}
  \subfigure[]{
    \label{CaPe} %% label for first subfigure
    \includegraphics[width=1.5in]{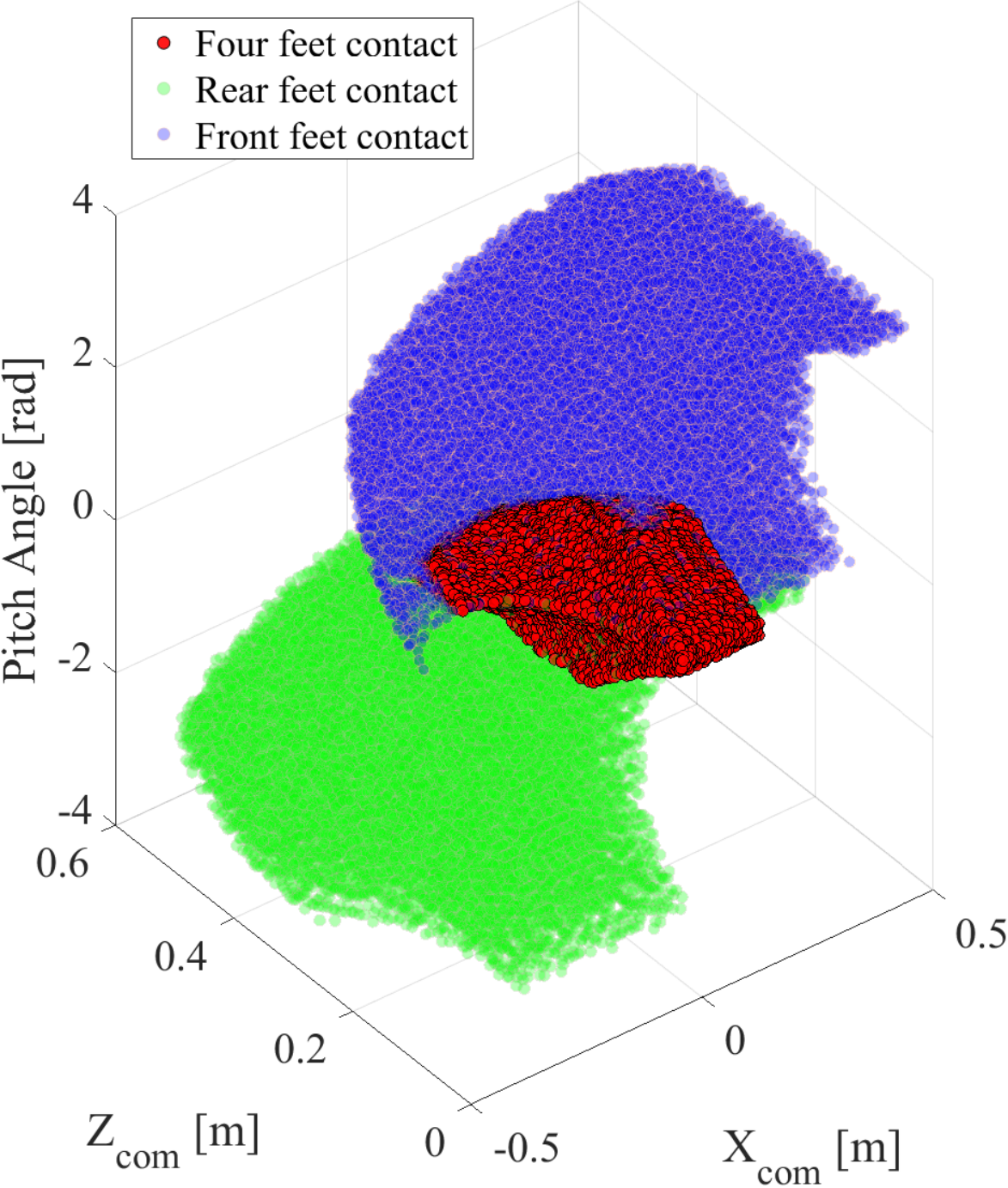}}
  \caption{The 3-dimensional (3D) configuration space for different jumping tasks and contact modes. (a) Front jump configuration space with the front, rear, and four feet contact modes. (b) Side jump configuration space with left, right, and four feet contact modes.}
  \label{C-Space} %% label for entire figure
  \vspace{-0.3cm}
\end{figure}

This section aims to build the configuration space (C-space)x. We introduce the kino-dynamic constraints including joint constraints, contact force constraints, and friction constraints \cite{Zt_2022} to generate the C-space, which makes this optimization problem lie in a much smaller searching region.
Inspired by Ding's work\cite{Ding_14}, we search for $\bm s_\Omega (\bm t_{\textrm{opt}})$ and $\bm t_{\textrm{opt}}$ of $\bm D_{\textrm{opt}}$ in two independent spaces, configuration space (C-space) $\bm {\Omega}_C \subset  \mathbb{R}^{3}$ and time-space (T-space) $\bm {\Omega}_T \subset \mathbb{R}^{3}$, respectively. The definition of $\bm {\Omega}_C$ and $\bm {\Omega}_T$ are as follows:
\begin{eqnarray}
\begin{aligned}
\bm {\Omega}_C:= \{\bm s_\Omega \in \mathbb{R}^{3} \ \arrowvert \ \bm q_{\textrm{min}}<\bm{q}(\bm {s_\Omega})<\bm q_{\textrm{max}}, \\
\bm{z}_{\textrm{hip}}(\bm {s_\Omega})>\bm z_{\textrm{min}},\\
\bm{z}_{\textrm{knee}}(\bm {s_\Omega})>\bm z_{\textrm{min}}\},\\
\bm {\Omega}_T:= \{\bm t_{\textrm{opt}} \in \mathbb{R}^{3} \ \arrowvert \ 0.1<\bm t_{\textrm{opt}}<0.5 \},
\end{aligned}
\end{eqnarray}
where $\bm {\Omega}_C$ is a set of robot's configurations $\bm s_\Omega$ in different jumping tasks and phases w.r.t. world frame, which satisfies joint angle and joint position constraints. The constraint of joint position ($z_{\textrm{hip}}$ and $z_{\textrm{knee}}$) means that the hip joint and knee joint should not be in contact with the ground during the jump. $\bm {\Omega}_T$ is the time range of four feet contacts, two feet contacts, and flight jumping phases that are manually selected. 
For $\bm {\Omega}_C$, due to the complex relationship between $\bm s_\Omega$ and $\bm q$, $\bm z_{\textrm{hip}}$, and $\bm z_{\textrm{knee}}$ the shape of $\bm {\Omega}_C$ is difficult to describe with analytical formulas. The value range of the three elements in $\bm s_\Omega$ depends on hardware limitations. Then we split each value range into 50 equal parts to get 125000 points. Finally, the shape of $\bm {\Omega}_C$ can be obtained (see Fig. \ref{C-Space}) by removing the points that do not satisfy the constraints of joint angles and joint positions. Therefore, for different jumping tasks and feet contact modes, the DLC algorithm can directly optimize $\bm s_\Omega$ in the corresponding $\bm {\Omega}_C$ to speed up the progress of finding local optimal $\bm s_\Omega$.
\vspace{-0.2cm}

\begin{figure}[!h]
  \centering
 \subfigure[]{
    \label{four_feet_pre} %% label for second subfigure
    \includegraphics[width=1.5in]{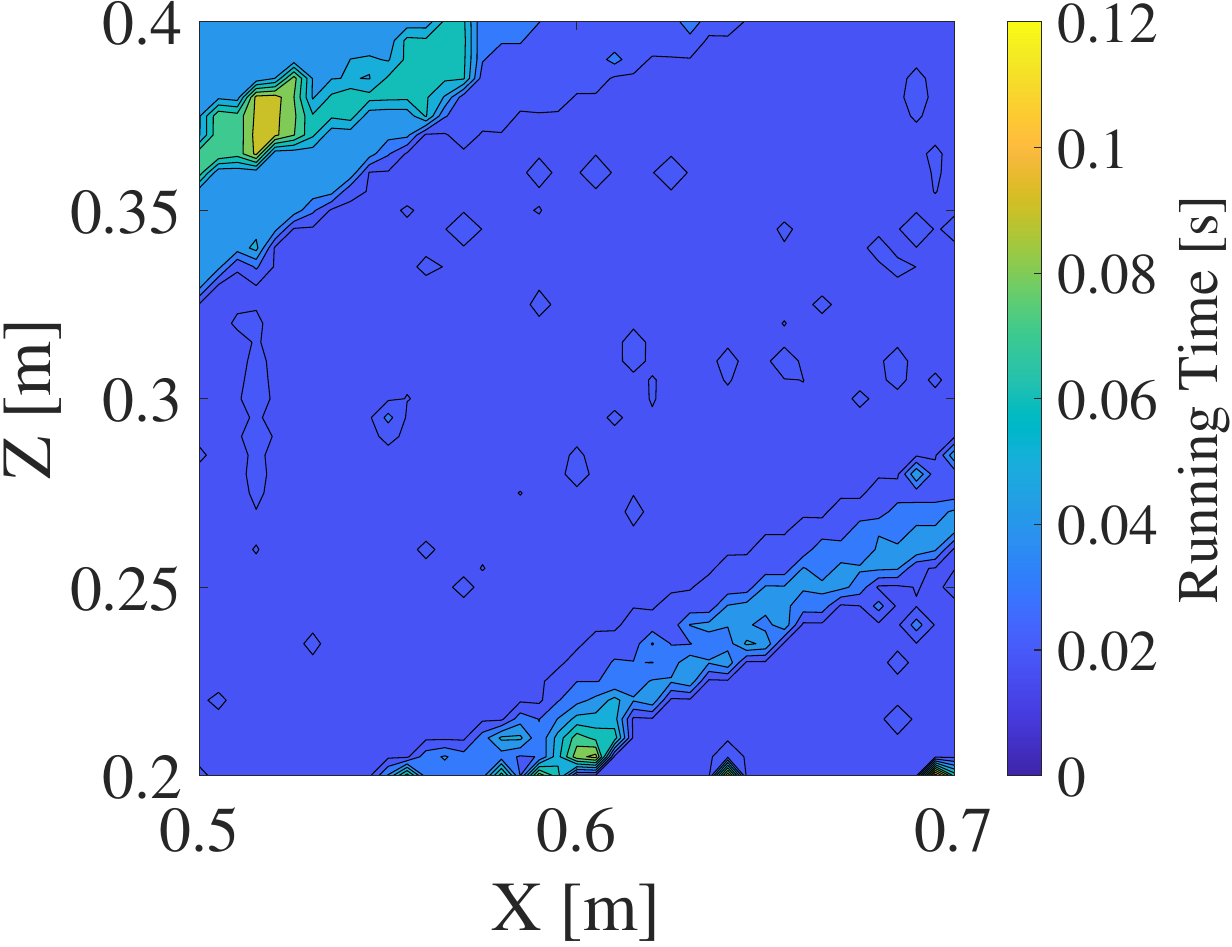}}
  \subfigure[]{
    \label{With_latin} %% label for first subfigure
    \includegraphics[width=1.7in]{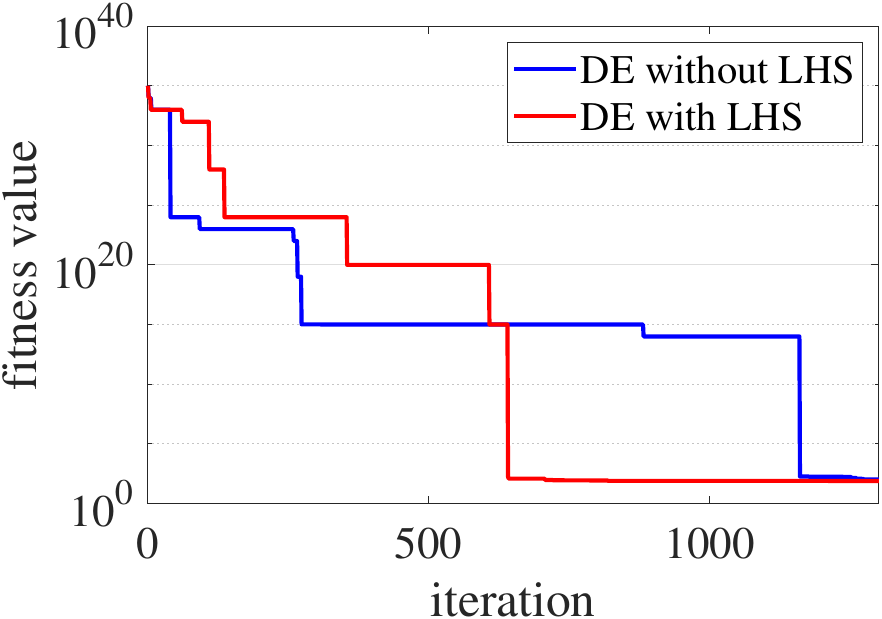}}
  \caption{The optimization time with the Pre-motion Library of four-leg front jump motion and convergence comparison of DE algorithm with and without LHS. (a) The DLC algorithm running time of the four contact front jumping task with $\bm D_{res}^*$ of ${\bm{s}_\Omega^*} = [0.6, 0.2, 0]$ in Pre-motion Library. The high-level information $\bm{s_t} = [x_c, z_c, 0]$, where $x_c \in [0.5, 0.7]\ m, z_c \in [0.2, 0.4]\ m$ and a sampling point was taken every 0.005 (m) in the two directions. (b) DE algorithm with and without LHS.}
  % \label{C-Space} %% label for entire figure
  % \vspace{-0.7cm}
\end{figure}
\begin{figure}[!h]
  \centering
\includegraphics[width=3.2in]{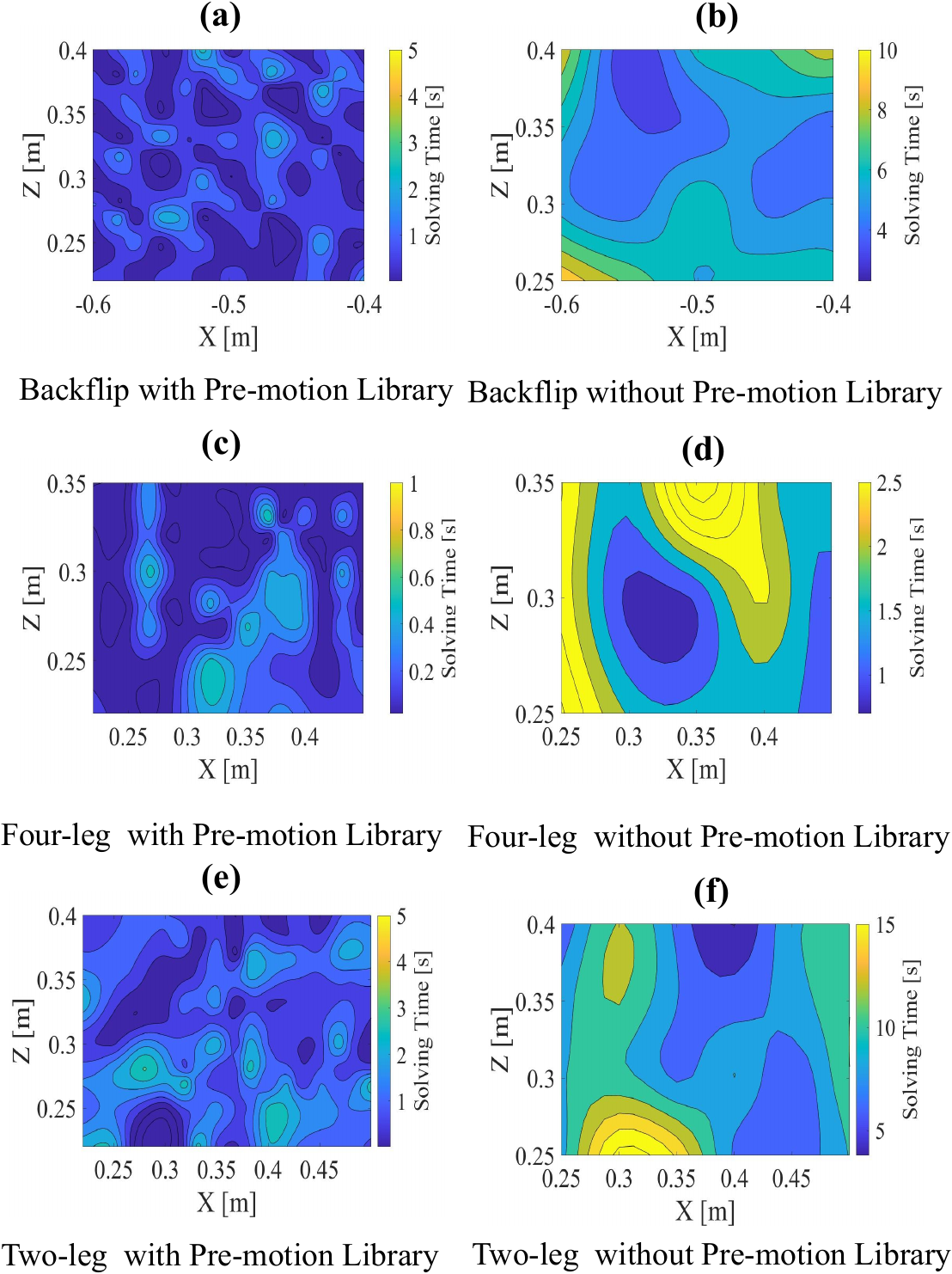}
  \caption{The DLC framework solving time on four-leg jump, two-leg jump, and back-flip with or without Pre-motion library. The solving time has random noise ($\pm 0.05$ (m)) at desired $\bm{s}_{\Omega}$. (a), (c) and (e) are the back-flip, four-leg jumping, and two-leg jumping solution times with the Pre-motion Library. (b), (d) and (f) are the solution time without the Pre-motion Library of those three jump motions.}
  \label{pre_motion_tree_time} %% label for entire figure
  % \vspace{-0.5cm}
\end{figure}
% \begin{figure}[!h]
%   \centering
%  \subfigure[]{
%     \label{w_p_b} %% label for second subfigure
%     \includegraphics[width=1.6in]{figs/pre_motion_lib_time/with_p_motion_back_flip.jpg}}
%   \subfigure[]{
%     \label{wo_p_b} %% label for first subfigure
%     \includegraphics[width=1.6in]{figs/pre_motion_lib_time/without_p_motion_backflip.jpg}}
%      \subfigure[]{
%     \label{w_p_f} %% label for second subfigure
%     \includegraphics[width=1.6in]{figs/pre_motion_lib_time/with_p_motion_four_jump.jpg}}
%   \subfigure[]{
%     \label{wo_p_f} %% label for first subfigure
%     \includegraphics[width=1.6in]{figs/pre_motion_lib_time/without_p_motion_four_jump.jpg}}
%      \subfigure[]{
%     \label{w_p_t} %% label for second subfigure
%     \includegraphics[width=1.6in]{figs/pre_motion_lib_time/with_p_motion_two_jump.jpg}}
%   \subfigure[]{
%     \label{wo_p_t} %% label for first subfigure
%     \includegraphics[width=1.6in]{figs/pre_motion_lib_time/without_p_motion_two_jump.jpg}}
%   \caption{The DLC framework solving time on four-leg jump, two-leg jump, and back-flip with or without Pre-motion library. The solving time has random noise ($\pm 0.05$ (m)) at desired $\bm{s}_{\Omega}$. (a), (c) and (e) are the back-flip, four-leg jumping, and two-leg jumping solution times with the Pre-motion Library. (b), (d) and (f) are the solution time without the Pre-motion library of those three jump motions.}
%   \label{pre_motion_tree_time} %% label for entire figure
%   \vspace{-0.5cm}
% \end{figure}

\begin{center}
% \vspace{1.2cm}
\begin{figure*}[!h]
\centering
\includegraphics[width=6.5in]{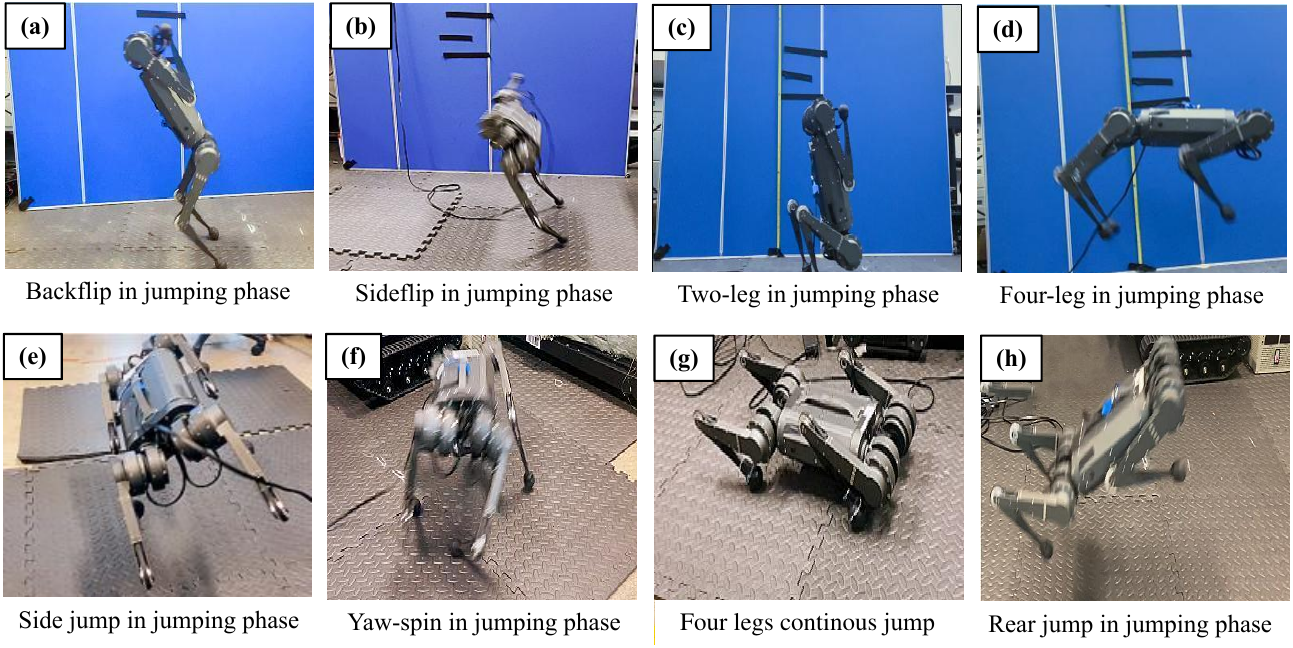}
\vspace{-0.2cm}
\caption{Different jumping motions experiments to validate the proposed DLC framework. (a) Back-flip with desired pitch angle at $\theta=-2\pi$(rad). (b) Side-flip with desired roll angle at $\varphi=-2\pi$ (rad). (c) Two-leg vertical jump with desired pitch angle at $\theta=-\frac{\pi}{2}$ (rad). (d) Four-leg vertical jump with desired height of 0.8 (m). (e) Side jump with a desired distance of 0.3 (m). (f) Yaw-spinning with desired yaw angle $\pi$ (rad). (g) Four legs continuous jump. (h) Rear jump with desired distance 0.3 (m).} \label{back_flip_snp}
\vspace{-0.4cm}
\end{figure*}
\end{center}

\begin{figure*}[!h]
  \centering
% \hspace{-2mm}
 \subfigure[]{
    \label{CaRe1} %% label for second subfigure
    \includegraphics[width=0.238\textwidth]{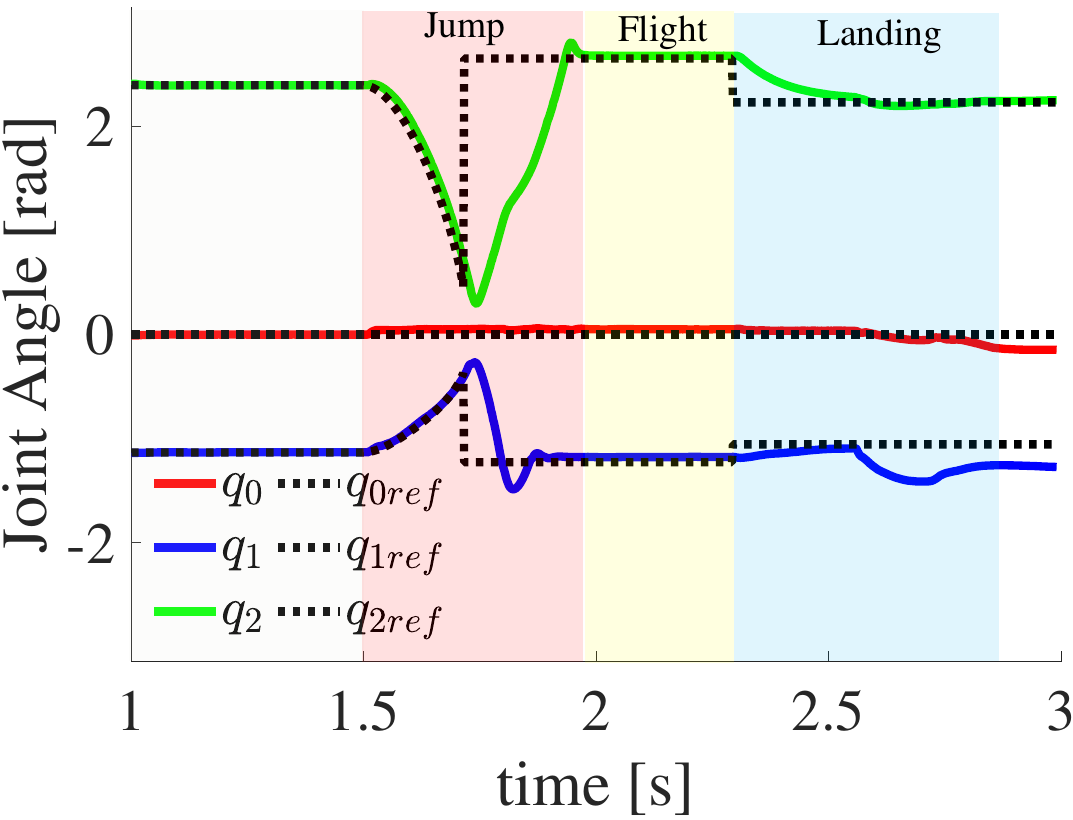}}
  \subfigure[]{
    \label{CaPe2} %% label for first subfigure
    \includegraphics[width=0.238\textwidth]{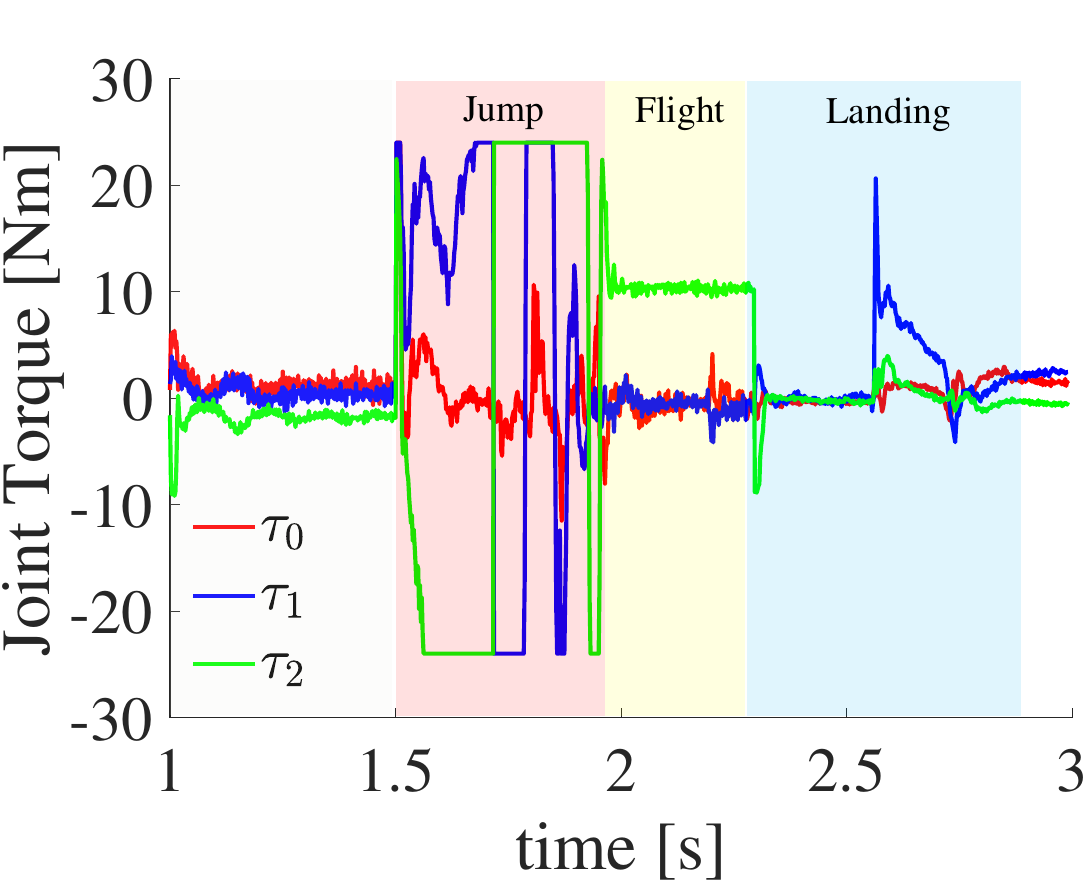}}
    \subfigure[]{
    \label{CaRe3} %% label for second subfigure
    \includegraphics[width=0.238\textwidth]{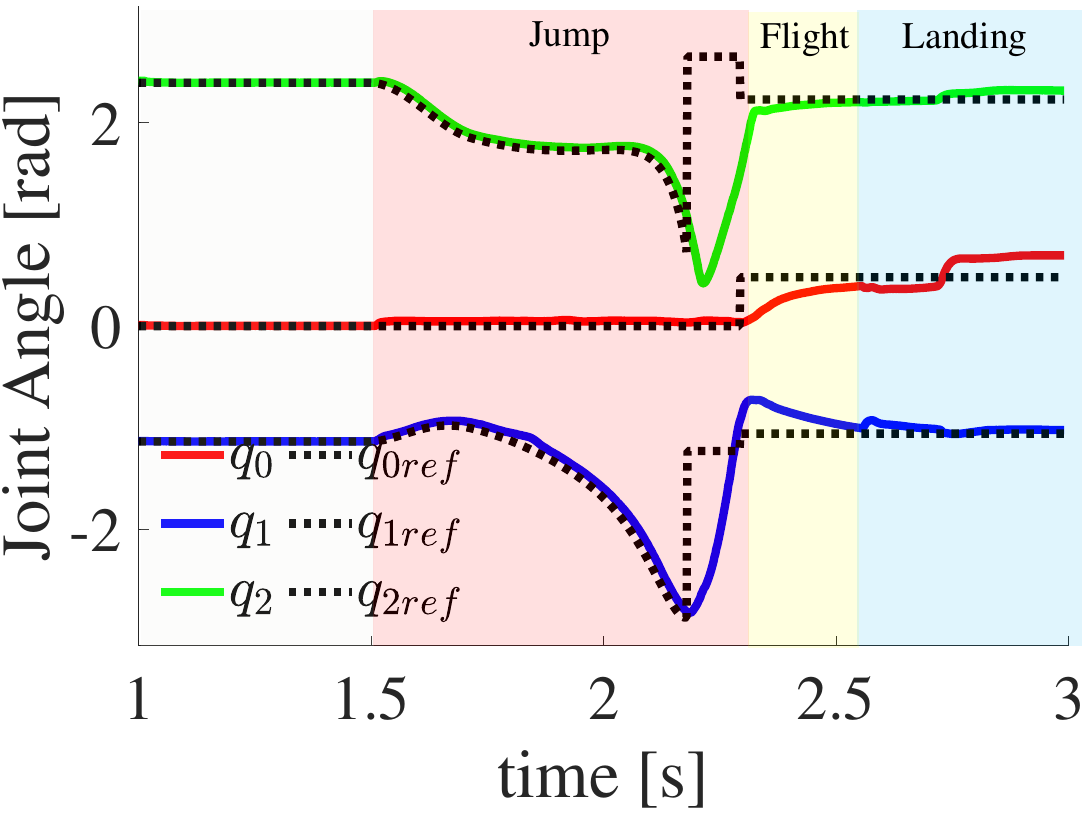}}\
  \subfigure[]{
    \label{CaPe4} %% label for first subfigure
    \includegraphics[width=0.238\textwidth]{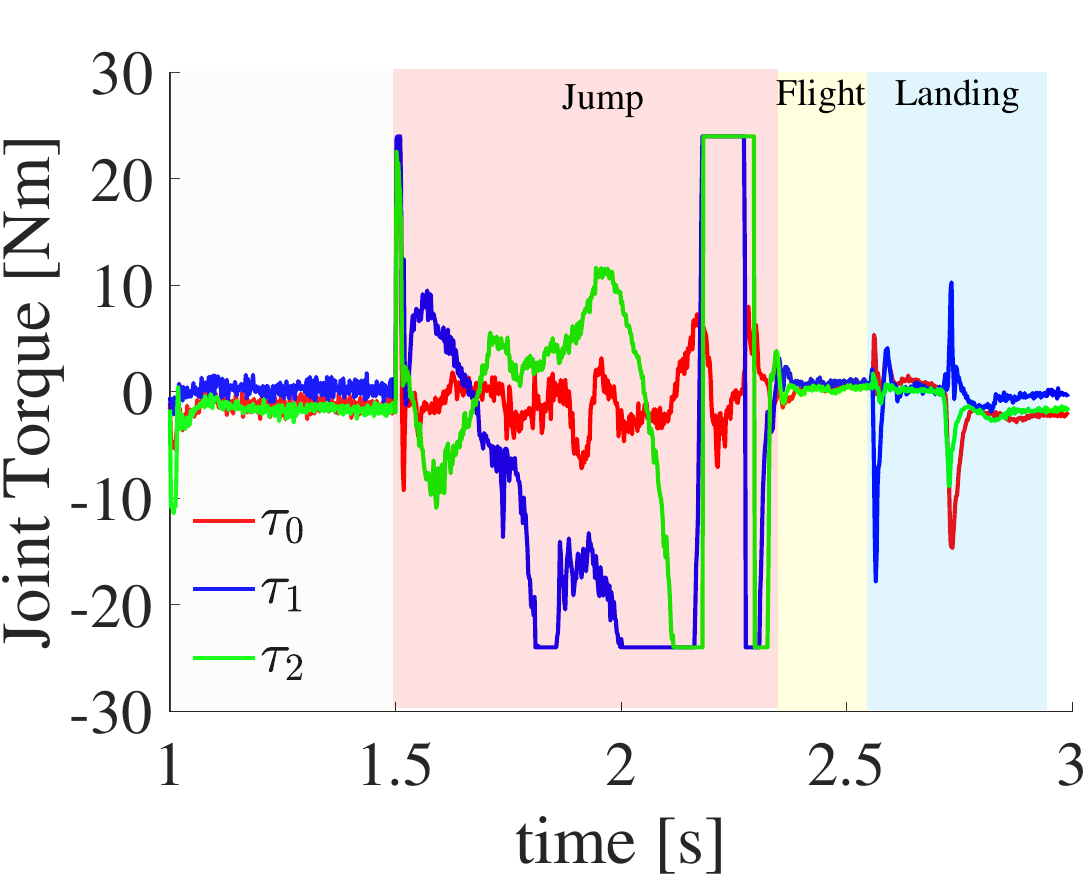}}
\vspace{-3mm}
  \caption{Joint angle and torque of the back-flip: leg 0 and leg 3 joint angles are shown in (a) and (c). (b) and (d) represent the joint torques of leg 0 and leg 3. Different color shadows of the figure show the different jumping phases of the different legs.}
  \label{back_flip_data} %% label for entire figure
  \vspace{-0.5cm}
\end{figure*}

\vspace{-0.5cm}
\subsection{Pre-motion Library}
This section aims to establish an offline library about a set of the 12 local optimal optimization variables ($\bm D_{\textrm{res}}$) called the Pre-motion Library to accelerate online optimization. 
% According to the experimental results, the online optimization time can be reduced from $\sim3\ (s)$ to $\sim0.14\ (s)$ (see Fig. \ref{four_feet_pre}) in a laptop (Intel(R) Core(TM) i7-9750H) with Pre-motion Library for four-leg jump (ref to Table. \ref{tab:avg_time}).

The central idea for building the Pre-motion Library is to generate a set of $\bm D_{\textrm{res}}$ offline with 12 local optimal optimization variables ($\bm{D}^*_{\textrm{res}}$). According to the fixed step size ($\sim0.05\ m$), $\bm{s}_{\Omega}$ is uniformly divided to obtain the target state of the robot. Then, the obtained target robot states are input into the DLC framework. The corresponding $\bm{D}^*_{\textrm{res}}$ are saved and collected to form the Pre-motion Library. The library comprises 567 $\bm{D}_{\textrm{res}}$ representing various kinds of jumping motion (e.g., front/rear/side, backflip, side flip, two/four-leg jump). When re-planning is required, pre-calculated optimization variables can be shared with new evolution as a warm start.
% Furthermore, in addition to the $D_{\textrm{res}}$, each record also includes the desired position and orientation of the robot CoM. There is also geometric information about the obstacles, especially the window-shaped obstacles are represented with the above and below rectangles.

An index file (Yaml-file) maintains all of the CoM's pre-motion trajectories. In addition, the Pre-motion file is $\sim30$ Megabytes in size (MB). It will be loaded into memory at the start of the controller's engine.
The desired trajectory from the library is based on the minimum Euclidean distance with a specified threshold (0.05 (m)), the input is the high-level information $\bm s_{\Omega}$. 
% After that, the $D_{\textrm{res}}$ will be used to initialize optimization variables instead of Latin hypercube sampling values (see Algorithm \ref{algo_de}) in the second-level DLC algorithm. 
% The final $\bm D_{\textrm{res}}$ as warm-start is then passed to the robot dynamics model in order to produce trajectories (see Fig. \ref{jump_framework}). If no trajectory meets all requirements, the selector will return false, and the framework will only employ online DLC optimization without using the trajectory in the library.
\begin{table}[H]
\center
\setlength{\tabcolsep}{1mm}{
\caption{Average solve times for different jump motion type}\label{tab:avg_time}
\begin{tabular}{c|c|c|c}
Jump Motion & Four-contact  & Back-flip &Two-contact \\ \hline \hline 
With Pre-motion Library & 0.14 (s) & 0.79 (s)&0.657 (s)\\
Without Pre-motion Library &  2.19 (s)& 5.7 (s)&9.11 (s)\\
Offline DE\cite{Zt_2022} &65 (s)& 167 (s)&266 (s)\\ \hline \hline 
\end{tabular}}

% \caption*{Rated Speed and Rated Current is 1250RPM and 15A}
\vspace{-0.3cm}
\end{table}

\subsection{Online DLC Algorithm}
After the C-Space, Pre-motion Library, and optimization variables are proposed, then the online DLC framework can be introduced. 

The online DLC algorithm utilizes the prioritization fitness function (see (\ref{optimization_formula})) to search for solutions in C-space. 
% The Pre-motion Library to make the online DLC algorithm converge faster. Online DLC optimization generates $\bm D_{\textrm{res}}$ by combining high-level information about the environment and ${{\bm D}_{\textrm{res}}^*}$ from Pre-motion Library and transmitting it to the robot dynamics model to generate trajectories. 
% In order to enable crossover and mutation operations in the differential evolution (DE) method, the optimization parameter vector $\bm D_{\textrm{opt}}$ is translated to a unit vector ${\bm s_m}$ (Algorithm. \ref{algo_de}).
We introduce ${\bm \Omega_{C}}$ and ${\bm \Omega_{T}}$ into the searching space ${\bm \Omega_{s}}$ to limit the mutation region. Then, in contrast to the conventional DE algorithm's random population initiation, we use Latin hypercube sampling (LHS) to produce a more uniform initial population distribution\cite{latin_20}, which improves algorithm iteration convergence speed (see Fig. \ref{With_latin}).  For the objective robot state $\bm{s}_t$ whose Euclidean distance to $\bm{D}_\textit{res}^*(t_3)$ is less than the threshold $\varepsilon$, it is obtained from the Pre-motion library, otherwise, it continues to use the LHS for initialization.

The details of the DLC algorithm are shown in algorithm \ref{algo_de}, where ${\bm s_t} \in {\mathbb{R}^3}$ is the desired position and Euler angular of the CoM and ${\bm O_k} \in {\mathbb{R}^{12}}$ is the location of obstacles coming from the high-level information in our framework. ${\textrm{Maxgen}}$ and ${NP}$ represent the maximum generation and population numbers, respectively. ${r}$ is the neighborhood radius of ${ {\bm D}_{\textrm{res}}^*}$. ${\varepsilon}$ is the fitness value at which the algorithm stops and returns $\bm D_{\textrm{res}}$, which is usually less than $\beta$. $g$ is the number of the DLC generations, $M(\cdot)$, $C(\cdot)$, and $\textrm{LHS}(\cdot)$ present the mutation, crossover, and Latin hypercube sampling functions. ${\bm U_m(g)}$ is the unit vector w.r.t. optimization parameters.

\section{Implementation Details and Experiments}
This section's primary objective is to experimentally verify the efficacy and adaptability of the proposed framework via various jumping motion types with the open-source MIT Mini-Cheetah\cite{Katz_19}. The jumping controller uses a joint-level PD controller with DLC generated torque. And a first-order low-pass filter for $\bm{q}$ and $\bm{\tau}$ is used for the landing controller. In order to protect the mechanical components of the robot, positive flexible landing control is necessary, here we employ relatively small PD gains\cite{Zt_2022}. 
% \subsection{Software Implementation}
% Three steps should be implemented in this part. First, construct the Pre-motion library offline on our Laptop, not Robot on-board computer. We first use Matlab to create the polynomial equations (\ref{coefficient_front}), and then we utilize the solution function built into Matlab to determine the polynomial coefficients vector (\ref{coefficient_d}) w.r.t. time.

% With the coefficients vector, we can plan the GRFs with those equations and then put those in an optimization iterative loop to generate the $\bm D_{\textrm{res}}$ with constraints and C-Space. As our database, we then store all $\bm D_{\textrm{res}}$ in key and value pairs in a YAML file. The online DLC optimization algorithm is first validated in the simulation platform. 

% At last, we transfer the library and online DLC optimization to the robot. There are two techniques for optimization: one employs the onboard computer (Intel ATOM x5-Z8350) directly, and the other utilizes the host computer (Intel(R) Core(TM) i7-9750H) to optimize and sends the results to the robot through UDP protocol.
% \subsection{Experiments}
% This section displays various jumping motions on an open-source Mini-Cheetah implementing our suggested framework\cite{Katz_19}. 
To study the solving efficiency of the proposed framework, we conduct experiments on the trajectory of optimization on both the simulation and the real robot. 
We do not just repeat the verification of the feasible solutions in the jump library; rather, we conducted a small-scale randomization ($\pm 0.05 (m)$ of the target position) of each offline-obtained feasible solution to evaluate the algorithm's adaptability. The optimization time for general motions (without touching the C-space boundary and hardware limits) is often less than 0.3 (s) (see Fig. \ref{pre_motion_tree_time}) with the Pre-motion Library. Still, it will take about 3-9 seconds to optimize actions with extreme boundaries (such as the highest vertical jump or highest double-leg jump height). The average solving time is shown in the Table.~\ref{tab:avg_time}. Using back-flip as an illustration, the average optimization time spans from 197s to 0.79s, while the solution speed is approximately 200 times faster. The DE algorithm with LHS typically requires fewer iterations than the conventional initial population technique (see Fig. \ref{With_latin}).
Our experiments organize into five categories: jumping motions, flipping motions, flipping from a platform, yaw-spin jumps, and vertical jumps. 
% The proposed framework also gave the robot the ability to perfect repeated jumps. 
The supplementary video contains demonstrations of the experiments.

In the flipping motion studies, our framework is employed to validate the back-flip, left-flip, and flip from a platform (see Fig. \ref{back_flip_snp} and Fig. \ref{problemIllustration}). The robot can perform a backflip from a 34-centimeter-high platform and land safely. For the second DLC technique, the offline-generated library is maintained onboard. The experimental data for back-flip jumping is depicted in Fig. \ref{back_flip_data}. 
% When a robot performs a jumping motion, there is often a large movement in the joint angle, as demonstrated by the joint angles and torques.
The experimental data shows that the torque is restricted at the maximal joint torque of 24 (Nm), showing that the robot requires a great deal of energy to leave the ground. Initially, we optimize the jumps using the robot's own computer (Intel ATOM x5-Z8350); however, with pre-motion, it takes the robot around $\sim3s$ or even large to optimize due to computational restrictions. Hence, a remote computer optimized online and sent the robot's trajectory through UDP.
\section{CONCLUSIONS}
In this paper, a novel online evolutionary-based time-friendly optimization motion planning framework has been proposed for quadruped jumping. Experiments show that an evolutionary-based method can be an alternative approach to solving the complicated motion planning problems of legged robots. Optimization variables transformation and C-space compress the DE searching region, and Latin hypercube sampling gives a more uniform initial population with limit points, which enhances the ability of the DE algorithm to escape from the local minimum. Those three core contributions can give a very obvious improvement in the convergence speed of the evolutionary algorithm.
 In particular, a small-scale random perturbation to the feasible solution in the pre-motion library is nevertheless capable of ensuring the approximate convergence speed. At the same time, the feasible solution in Pre-motion Library as a warm-start can significantly boost the framework optimization progress. Experimental results indicate a significant reduction in the convergence speed compared with our previous work\cite{Zt_2022}.
 
 Additionally, our framework prioritizes optimizing time consumption rather than landing precision of jumping. And, the optimization time for extreme cases which touch the C-space boundary (e.g., jumping to a high enough desk (0.4 (m)) or side-flipping over high enough obstacles) still needs to be shortened.

\addtolength{\textheight}{0cm}   % This command serves to balance the column lengths
                                  % on the last page of the document manually. It shortens
                                  % the textheight of the last page by a suitable amount.
                                  % This command does not take effect until the next page
                                  % so it should come on the page before the last. Make
                                  % sure that you do not shorten the textheight too much.

%%%%%%%%%%%%%%%%%%%%%%%%%%%%%%%%%%%%%%%%%%%%%%%%%%%%%%%%%%%%%%%%%%%%%%%%%%%%%%%%

%%%%%%%%%%%%%%%%%%%%%%%%%%%%%%%%%%%%%%%%%%%%%%%%%%%%%%%%%%%%%%%%%%%%%%%%%%%%%%%%

%%%%%%%%%%%%%%%%%%%%%%%%%%%%%%%%%%%%%%%%%%%%%%%%%%%%%%%%%%%%%%%%%%%%%%%%%%%%%%%%
% \section*{APPENDIX}

% Appendixes should appear before the acknowledgment.

% \section*{ACKNOWLEDGMENT}

% The preferred spelling of the word ÒacknowledgmentÓ in America is without an ÒeÓ after the ÒgÓ. Avoid the stilted expression, ÒOne of us (R. B. G.) thanks . . .Ó  Instead, try ÒR. B. G. thanksÓ. Put sponsor acknowledgments in the unnumbered footnote on the first page.

% %%%%%%%%%%%%%%%%%%%%%%%%%%%%%%%%%%%%%%%%%%%%%%%%%%%%%%%%%%%%%%%%%%%%%%%%%%%%%%%%

% References are important to the reader; therefore, each citation must be complete and correct. If at all possible, references should be commonly available publications.

\end{document}